\documentclass{article}

\usepackage{todonotes}

\newcommand{\node}[1]{\ensuremath{n}_{#1}}

\usepackage[preprint, nonatbib]{neurips_2019}
\usepackage{wrapfig}
\usepackage[utf8]{inputenc} 
\usepackage[T1]{fontenc}    
\usepackage{hyperref}       
\usepackage{url}            
\usepackage{booktabs}       
\usepackage{amsfonts}       
\usepackage{amsmath}
\usepackage{amssymb}
\usepackage{amsthm}
\newtheorem{asu}{Assumption}

\newtheorem{prp}{Proposition}
\newtheorem{dfn}{Definition}
\usepackage{xcolor}
\usepackage{float}
\usepackage{subfig}
\usepackage{algorithm}
\usepackage[noend]{algpseudocode}
\usepackage{graphicx}
\usepackage{graphics}
\usepackage{nicefrac}       
\usepackage{microtype}      
\usepackage{wrapfig,lipsum,booktabs}

\title{GLANCE: Global to Local Architecture-Neutral Concept-based Explanations}

\author{%
  Avinash Kori\\
  Department of Computing\\
  Imperial College London\\
  \texttt{a.kori21@ic.ac.uk} \\
  \And
  Ben Glocker\\
  Department of Computing\\
  Imperial College London\\
  \texttt{b.glocker@ic.ac.uk} \\
  \And 
  Francesca Toni\\
  Department of Computing\\
  Imperial College London\\
  \texttt{f.toni@ic.ac.uk} \\
}

\begin{document}

\maketitle

\begin{abstract}
Most of the current explainability techniques focus on capturing the importance of features in input space. 
However, given the complexity of models and data-generating processes, the resulting explanations are far from being `complete', in that they lack an indication of feature interactions and visualization of their `effect'.
In this work, we propose a novel twin-surrogate explainability framework to explain the decisions made by any CNN-based image classifier (irrespective of the architecture). 
For this, we first disentangle latent features from the classifier, followed by aligning these features to observed/human-defined `context' features.  
These aligned features form semantically meaningful concepts that are used for extracting a causal graph depicting the `perceived'  data-generating process, describing the inter- and intra-feature interactions between unobserved latent features and observed `context' features.
This causal graph serves as a global model from which local explanations of different forms can be extracted.
Specifically, we  provide a generator to visualize the `effect' of interactions among features in latent space and draw feature importance therefrom as local explanations.
Our framework utilizes adversarial knowledge distillation to faithfully learn a representation from the classifiers' latent space and use it for extracting visual explanations. 
We use the styleGAN-v2 architecture with an additional regularization term to enforce disentanglement and alignment.
We demonstrate and evaluate explanations obtained with our framework on Morpho-MNIST and on the FFHQ human faces dataset.
Our framework is available at \url{https://github.com/koriavinash1/GLANCE-Explanations}
\end{abstract}

\section{Introduction}

Deep learning models have emerged as powerful tools for solving complex problems in diverse domains in the past decade, and still, they are considered as \textit{black-boxes} due to their lack of interpretability.
At the same time, there is a consensus among researchers, ethicists,  policy makers and the public on the need for explainability of these models, especially in high-stake applications like bio-medicine and autonomous driving \cite{doshi2017role, kroll2015accountable}.
Explaining decisions made by deep learning classifiers can not only help us understand the underpinning mechanism but also uncover model biases \cite{kim2018interpretability}, which helps in better understanding the data-generating process \cite{narayanaswamy2020scientific}.
There are many different forms of explainability techniques, including feature attribution methods \cite{ribeiro2016should}, network dissection-based interpretability \cite{bau2017network}, mechanistic approaches for understanding neural networks \cite{olah2017feature, olah2020zoom}, and causal/counterfactual explanations \cite{sauer2021counterfactual, chang2018explaining, pawlowski2020dscm}. 
In this paper, we contribute to this landscape by defining a novel method for obtaining concept-based explanations.

Interpretability can be divided into two categories \cite{lipton2018mythos}: transparency and \emph{post-hoc explanations}; most of the above mentioned techniques fall under the latter category, as does our proposed framework.
Many existing frameworks for post-hoc explainability do not reflect concept-based thinking of the kind exhibited by humans \cite{armstrong1983some}, with a few recent exceptions. \cite{ghorbani2019towards} shows the existence of these concepts, while \cite{kori2020abstracting} uses the idea of both existence and interaction between concepts to generate explanations.
Our proposed framework generates concept-based explanations using unobserved latent and observed context features as concepts and identifying interactions between them.

\begin{figure}[t]
    \centering
    \includegraphics[width=1\textwidth]{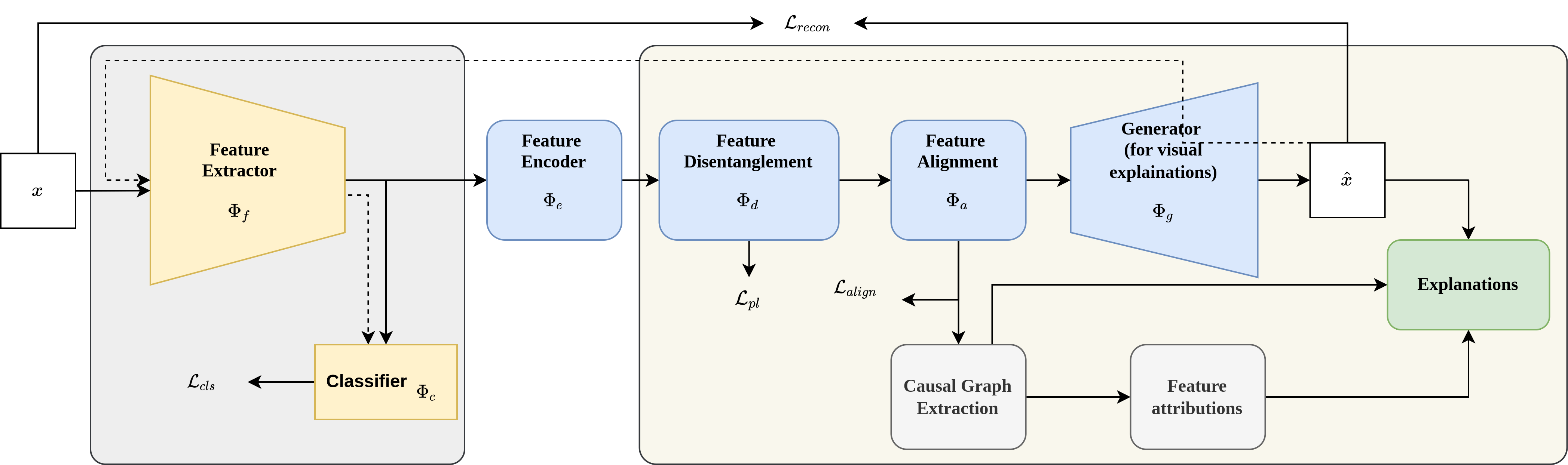}
    \caption{Overview of the proposed framework, in which 
    the feature extractor ($\Phi_f$) and the feature classifier ($\Phi_c$) are blocks 
    of a trained, given classifier model (
    $\Phi_C$).
    The feature disentanglement ($\Phi_d$), feature alignment
    ($\Phi_a$), and generator
    ($\Phi_g$)  blocks are part of our proposed twin-surrogate model. The causal graph extraction and feature attribution blocks, together with the generator, provide  our explanations.}
    \label{fig:overview}
    \vspace{-15pt}
\end{figure}

Among the different forms of explanations, counterfactual explanations are recently gaining attention \cite{sauer2021counterfactual, chang2018explaining, nemirovsky2020countergan, lang2021explaining}.
These help us analyse a classifier by constructing hypothetical scenarios and observing classifier predictions.
At the same time, the language of causality is advocated as a precise and powerful way of extracting explanations \cite{o2020generative}.
Counterfactual explanations can be drawn by intervening on the set of features in the data-generating process to construct hypothetical scenarios.
The effectiveness of counterfactual explanations solely depends on an intuitive difference between original and intervened data. 
In this work, we focus on generating \emph{causal graphs} using unobserved latent features which may or may not be human understandable and 
observed {context features} to model the data-generating process as perceived by the underlying model. The causal graph then serves as the basis for our explanations.

Methodologically, we use a twin-surrogate model based on disentanglement, alignment, and generator blocks as 
overviewed in Figure \ref{fig:overview}.
The generator is implemented as an extension of styleGAN-v2 \cite{karras2020analyzing} augmented to incorporate additional blocks for disentanglement and alignment.
We explore the latent space for causal discovery by following an intervention-based method, considering the trained twin-surrogate as an oracle to be consulted on the effects of interventions. 
The generator is used to extract counterfactual visual explanations to understand the `effect' of feature behaviour and interactions in causal discoveries.

Our goal in this work is to explain what a given CNN-based classifier learns, rather than explaining the true label.
We achieve this by learning a `pseudo' data-generating process along with a feature interaction graph,\footnote{By \textit{pseudo} we mean a data-generating process as perceived by the given classifier model.}  which serves as a \emph{global} ground for extracting our \emph{local} explanations (for given inputs).
These feature attribution based explanations respect the feature interactions in the extracted graph. The global and local explanations can be seen as two layers in a hierarchy, with the local explanations providing finer-grained information about the weight of features and their influence on the classifier's prediction, while reflecting the interactions in the global graph.
We  demonstrate and evaluate the effectiveness of our framework on Morpho-MNIST \cite{castro2019morpho}, a synthetic dataset based on MNIST \cite{deng2012mnist}, and the FFHQ dataset \cite{karras2019style}, a dataset with high quality human faces and attributes corresponding to facial features.
Overall, our contribution in this work is threefold:
\begin{itemize}
\item \textbf{Twin-surrogate model \ref{subsec:surrogate}:} we propose a novel twin-surrogate model which extends on the styleGANv2 framework for disentangling and aligning latent features for generating the classifier perceived data-generation process.
\item \textbf{Causal graph extraction \ref{subsec:graphs}:} we formalize a method to facilitate causal discovery of feature interactions among unobserved latent features and context features (focusing on causal structure rather than functional mechanisms). 
\item \textbf{Explanations \ref{subsec:exp}:} we propose a novel form of explanation that follows hierarchical steps of (i) global graph generation, capturing causal relationships as perceived by the model, and (ii) local feature attributions for a given input image along with a way to visualise and analyse feature interactions via counterfactuals.
\end{itemize}


\section{Related work}
Our method falls within an active field of research on post-hoc explainability for deep learning models.
Most post-hoc explainability methods can be categorized as (local or global) feature attribution-based or counterfactual explanations. 
Both feature attribution \cite{lime, SHAP} and counterfactual explanations \cite{goyal2019counterfactual, denton2019detecting} have proved to be useful methods to interpret the reasons for models decisions.
Feature attribution mainly focuses on estimating the importance weight for input features, indirectly indicating how features influence final decisions. 
In the case of images, features correspond to parts or patches of images responsible for the classifier's decisions \cite{lime}.
On the other hand, counterfactual explanations are influenced mainly by hypothetical \emph{``what-if''} scenarios. 
In case of generating counterfactual explanations for images,
black-box models are usually explained via \emph{twin-surrogate} models to provide visual explanations with desired latent properties \cite{dhurandhar2018explanations, denton2019detecting, goyal2019counterfactual, joshi2019towards, razavi2019generating, lang2021explaining}.
Most of these methods train generators from scratch, leading to explanations that are more \emph{faithful} to the given dataset than to the trained classifier. 
Some of these methods aim to generate samples that affect the classifier's decision \cite{dhurandhar2018explanations, joshi2019towards}, while others work on changing the latent space and observing the classifier’s decision change \cite{denton2019detecting}. 
The main focus in the case of twin-surrogate model-based explanations is to extract disentangled representations \cite{higgins2016beta}.
As there exist infinitely many possibilities to disentangle features, it has been shown that disentanglement without supervision is a challenging problem \cite{locatello2019challenging}.
Recent findings suggest that limited supervision can restrict the search space and 
can be used on a subset of latent features to optimize them to align towards desired properties \cite{locatello2020weakly}. 

Our work focuses on both feature attribution and the use of twin-surrogate models to derive faithful explanations.
Instead of training twin-surrogate models from data, we distill knowledge from the pre-trained classifier, making use of a generator to depict the classifier's perceived data-generating process.
We use the latent features from a the distilled model to determine the feature interactions, which is used to obtain counterfactual visual explanations and feature attributions indicating the contribution of each latent feature in towards the classifiers output. 

\section{Methods} 


In this section we describe the methods underpinning our explainability framework, aiming to demystify a given classifier by distilling it's knowledge into a twin-surrogate model, first learning to disentangle and align features, followed by causal discovery and generation of visual explanations.
The proposed method consists of five building blocks, as illustrated in Figure ~\ref{fig:overview}; the individual blocks are responsible for: (i) learning a disentangled representation, (ii) aligning disentangled features to observed context features, (iii) learning a generative decoder for visual explanations, (iv) constructing a causal graph, and, finally (v) deriving explanations.
Next, we explain each step in more detail.

\subsection{Preliminaries and notations}
\label{subsection:prelims_notations}
Let $\mathcal{D} \subseteq \mathcal{X} \times \mathcal{Y}$ be the dataset, such that elements of $\mathcal{X}$ are in  $\mathbb{R}^{s \times s}$ and $\mathcal{Y} 
= \{1, \ldots, N\}$, where $s \times s$ and $N$ correspond to the dimension of an input image and total number of classes, respectively.

Now, we define notations used to describe all the model components shown in Figure \ref{fig:overview}: (i) the pre-trained classifier is denoted by $\Phi_C: \mathcal{X} \rightarrow \mathcal{Y}$; (ii) the encoder block maps the classifiers' latent space to the required dimension\, denoted as $\Phi_e:\mathcal{E} \rightarrow \mathcal{E}'$, where elements of  $\mathcal{E}$ are in $\mathbb{R}^l$ and elements of $\mathcal{E}'$ are  in $\mathbb{R}^m$, with $m<l$  and, typically, $l,m \geq N$ - here $l$ corresponds to the dimension of the latent vectors and $m$ corresponds to; (iii) the disentanglement block is denoted by $\Phi_d: \mathcal{E}' \rightarrow \mathcal{E}'$ -  this block also serves as a modulator layer as proposed in styleGAN-v2 \cite{karras2020analyzing}; (iv) the alignment block helps in mapping the disentangled latent space to a human-understandable latent space with the help of observed context features, denoted as $\Phi_a: \mathcal{E}' \rightarrow \mathcal{E}'$; and (v) the generator block is used to construct visual explanations from latent space, denoted by $\Phi_g: \mathcal{E}' \rightarrow \mathcal{X}$.

\begin{asu} 
\label{assumption:classifier_decomposition}
We restrict 
our analysis 
to
classifiers that can be decomposed as $\Phi_C = \Phi_c \circ \Phi_f$, where feature extractor $\Phi_f:\mathcal{X} \rightarrow \mathcal{E}$ maps input images to latent vectors 
and feature classifier $\Phi_c:\mathcal{E} \rightarrow \mathcal{Y}$ maps the latent space to class labels.
\end{asu}

\textit{Remark}: This assumption requires the last block or layer of the classifier to map from embedding space to output space linearly; this is crucial because our methods use the embedding space to explain the decision made by the classifier. The assumption holds for many state-of-the-art classifiers.

\subsection{Twin-surrogate model}
\label{subsec:surrogate}
Our explainability framework involves learning a twin-surrogate model with a definite set of properties. 
The main properties of a twin-surrogate model that we consider are (i) feature disentanglement, (ii) feature alignment, and (iii) an ability to generate visual explanations.
Because feature disentanglement helps us construct a set of independent features responsible for the data-generating process, the feature alignment property transforms those obtained features into semantically meaningful features, and the generator helps us to obtain visual explanations, explaining these features.
Due to feature disentanglement and generator, our implicit choice of model reduced to variational auto-encoders \cite{kingma2019introduction} or generative models \cite{goodfellow2014generative}. 
We adopt the discoveries from \cite{locatello2019challenging}, which shows how variational and adversarial training encourages models to learn disentangled representations implicitly.
We further decided on the styleGAN-v2 \cite{karras2020analyzing} architecture because of its property to generate large scale images and disentangle latent space using path length regularization.
However, we noticed that some of the learned representations by styleGAN-v2 have a high correlation with others, and thus are not semantically meaningful; we experimentally demonstrate this in our analysis in appendix \ref{section:appendix}. 
To address this issues, we introduce an alignment block with additional loss to enforce independence and feature alignment. 

Let $\Phi_{sg}: \mathcal{E}' \rightarrow \mathcal{X}$ be 
the styleGAN-v2 framework, which, by construction, can be decomposed as $\Phi_{sg} = \Phi_{g} \circ \Phi_d$, where $\Phi_d$ and $\Phi_{g}$ correspond to the modulator and generator, respectively, as described in \cite{karras2020analyzing}. 
We define an alignment block $\Phi_{a}$, which maps disentangled features to semantically meaningful features with respect to observed context features, resulting in a decomposable generator model that can be described as $\Phi_{sg}' = \Phi_g \circ \Phi_a \circ \Phi_d$. 

Now, we describe the properties and assumptions considered in the construction of the alignment block.
Let $\mathcal{E}'_o, \mathcal{E'}_u$, correspond to \emph{context features} (i.e. observed, human-understandable features in the data-generating process)  and (unobserved) \emph{latent features} such that $\mathcal{E}'_o \subset \mathcal{E}'$, $\mathcal{E}_u \subset \mathcal{E}'$ and $\mathcal{E}'  = \mathcal{E}'_o \cup \mathcal{E}'_u$.
Typically $|\mathcal{E}'_o| < |\mathcal{E}'_u|$, namely the number of context features is less than the number of unobserved latent features: this makes our alignment task a problem of subspace alignment.



\begin{asu}
\label{assumption:depth_independence}
In the case of feature alignment, we assume that the given observed \emph{context features} follow a Directed Acyclic Graph (DAG) structure.
\end{asu}

\textit{Remark:} Basically our framework works on DAGs, namely we assume that features responsible for the data-generating process do not form any cycles or self-loops.

Based on $\mathcal{E}'$s information, we propose an alignment regularization term with the following properties:
\begin{itemize}
    \item \textbf{Regularization should involve subspace optimization}, which makes use of observed ground-truth context features; this also forces the model to encode \textit{relations} between context features and the morphology of an image.
    \item \textbf{Regularization should impose orthogonality} on features in $\mathcal{E}'_u$ among each other and also with respect to features in $\mathcal{E}'_o$; this helps in optimizing all the parameters in our alignment block, while just aligning a subset of features.
\end{itemize}

Let $\mathcal{C}$ correspond to ground-truth context features, the set of observed human-understandable features; by assumption \ref{assumption:depth_independence} all the elements in $\mathcal{C}$ form a DAG.


\begin{dfn}
The alignment of latent subspace to observed context features can be achieved by minimizing the L2 distance between the subspace of latent features and ground-truth context features.
This corresponds to $||\mathcal{E}'_u - \mathcal{C}||^2_2$, constrained on $z_i \perp z_j$, where $z_i, z_j \in \mathcal{E}'_u$ and $i \neq j$.
\end{dfn}

\textit{Remark:} The orthogonality constraint helps in limiting possible combinations of disentangled vectors to form aligned vectors. 
The alignment block is basically a linear transformation of disentangled features to match few observed context features, while  constraining on the rest.

To apply the orthogonality constraint, we first compute and track the running mean of eigenvectors, and condition the output of the alignment block to move close towards the mean eigenvectors.
We apply singular value decomposition (SVD) on matrix $M$, where $M$ is the submatrix of the batch output of the alignment block and $M \in \mathbb{R}^{b \times |\mathcal{E}'_o|}$, with $b$ being the  batch size used in training and $|\mathcal{E}'_o|$ corresponding to the number of observed context features and each row in $M \in \mathcal{E}'_u$. 
The aim of the alignment block is to force $\mathcal{E}'_u$ to align towards the mean eigenvectors of $M$. 
The SVD decomposition of $M$ can be described as $U\Sigma V^* = M$
, where $U, \Sigma, V^*$ correspond to left singular vectors, singular value matrix, and right singular vectors, respectively.
Eigenvectors of $M$ can be computed by simply multiplying left singular vectors with singular value matrix.
To control the maximum eigenvalue of unobserved latent features,  we normalize the eigenvector matrix ($U\Sigma$) with the Forbinious-norm $||.||_f$ of singular value matrix $\Sigma$.
Equation \ref{eqn:alignmentloss}, describes the proposed alignment loss mathematically, where $\hat{U\Sigma}$ corresponds to the running mean of an eigenvector of matrix $M$, $\lambda_{max}$ corresponds to a hyper-parameter to control the maximum eigenvalue of $M$, and $\alpha$ corresponds to weighage term of orthogonal conditioning.
The value of $\alpha$ is increased gradually from 0 to 1 with respect to training iterations (based on our experiments we found step based incremental function to work best). 

\begin{align}
    \begin{split}
      U \Sigma V^* &=  SVD(M), \; M \in \mathbb{R}^{b \times |\mathcal{E}'_u|}\\
    \mathcal{L}_{align} &= ||M - \mathcal{C}||^2_2 + \alpha \Big{\|}M - \frac{\lambda_{max} \hat{U\Sigma}}{||\Sigma||_f} \Big{\|}^2_2
    \end{split}
    \label{eqn:alignmentloss}
\end{align}

We use an adversarial training procedure to learn all encoder, disentanglements, alignment, and generative steps jointly. 
As opposed to randomly sampling a noise vector for generating images, we condition our noise distribution on the feature extractor from a trained classifier $z \sim \mathcal{N}(\Phi_f(x), 0.1)$, where $x \in \mathcal{X}$ and $\mathcal{N}(.)$ is normal distribution with given parameters. 
Due to this, our framework resembles an auto-encoder architecture with an additional discriminator and a fixed feature extractor; this requires an additional reconstruction loss term $\mathcal{L}_{recon}$ in training.
This is done to distill knowledge from the trained classifier to generate explanations that are meaningful and faithful to the classifier, while preserving the properties of styleGAN-v2. 
It is important to note that our generator here is reconstructing images as perceived by the classifier, not the original data.
The reconstructed images only contain features that the classifier sees as important in making its decision. 
The total generator loss is considered to be a linear combination of adversarial loss, reconstruction loss, path length, alignment loss, and cross-entropy loss, as follows:
$\mathcal{L}_{total} = \mathcal{L}_{adv} + \lambda_1 \mathcal{L}_{pl} + \lambda_2 \mathcal{L}_{align} + \lambda_3 \mathcal{L}_{recon} + \lambda_4 \mathcal{L}_{cls}$, where $\mathcal{L}_{cls}$ is cross-entropy loss applied on the classifier's prediction between original images and the classifier's perceived images, and the $\lambda_i$ are hyper-parameters to decide the weight for each loss component.
The total loss can also be described as $\mathcal{L}_{total} = \mathcal{L}_{styleGANv2} + \lambda_2 \mathcal{L}_{align} + \lambda_3 \mathcal{L}_{recon} + \lambda_4 \mathcal{L}_{cls}$.
We list all the hyper-parameters and other experimental specifications in appendix \ref{section:appendix}.

\subsection{Causal graph extraction}
\label{subsec:graphs}

We use diversity in latent space to aid an explanation for a classifier's decisions.
We argue that explaining via latent space features is more expressive compared to other feature attribution or saliency-based explanation methods.
As our explanations adopt both global and local perspectives, graph extraction is the central aspect of global explanation generation.
To extract a global graph, we take aligned latent features as the basic elements for constructing the pseudo data-generating process, which allows us to perform interventions and observe changes in the generated images.
As the generator learns the data distribution, it implicitly encodes functional mechanisms/relationships between variables responsible for generating the data.  
Here, we propose a method to extract feature interactions, without focusing on structural mechanisms.

\textbf{Causal discovery and graph extraction: }
We analyze the aligned features (ie., the output of the alignment block) to extract the learned relations among features represented as directed edges in a DAG.
After training, we can access the \textit{pseudo} data-generating process (generator model) as an oracle and perform controlled interventional queries. 
These amount to questions of the form \textit{``How would the generated image change if I change this particular feature?''}.
We determine the existence of directed edges between features by comparing original and intervened latent feature values.
Now we define some graph specific terms, which we use in our discovery step.

\begin{dfn}
Node $\node{i}$ and node $\node{j}$ in a DAG
are said to have a \emph{Direct Causal Path (DCP)} if there exists an edge between $\node{i}$ and $\node{j}$
(either $\node{i} \rightarrow \node{j}$ or $\node{j} \rightarrow \node{i}$), and are said to have an \emph{Indirect Causal Path (ICP)} if their exists a trail from $\node{i}$ to $\node{j}$ via a third node $\node{k}$ (either $\node{i} \rightarrow \node{k} \rightarrow \node{j}$ or $\node{j} \rightarrow \node{k} \rightarrow \node{i}$).
Finally, we define the \emph{edge-weight} for an edge between $\node{i}$ and $\node{j}$ as:
\begin{equation}
    EW(\node{i},\node{j}) \triangleq \mathbb{E}_{l=\Phi_a(\Phi_d(\Phi_e(z))), z \sim \mathcal{N}(\Phi_f(x), 0.1); x\sim \mathcal{X}} \frac{[\hat{l}_{j} - l_{j}]}{[\hat{l}_{i} - l_{i}]}
    \label{eqn:edgeweightage}
\end{equation}

where $l_i$, $\hat{l}_i$ indicate the $i^{th}$ element in vectors $l$,$\hat{l}$, respectively (with position $i$ corresponding to node $\node{i}$ in the graph, similarly for $j$)
and $\hat{l}$ is the \emph{intervened latent vector}, formally defined as  $\hat{l}= \Phi_a(\Phi_d(\Phi_e(z))); z \sim \mathcal{N}(\Phi_f(\Phi_g(l; do(l_i=I))), 0.1)$, for $I$ an intervention on $i$. 

\label{prp:causalstructure}
\end{dfn}

Procedurally, for causal discovery via interventional queries, we propose the following steps (for simplicity, below we equate nodes and positions in vectors): 

\begin{enumerate}
    \item We extract the aligned feature vector by passing a sampled feature vector through a composite function of encoder, disentanglement and alignment, formally described as $l = \Phi_a (\Phi_d ( \Phi_e (z))); z \sim \mathcal{N}(\Phi_f (
    x), 0.1)$, where $
    x\sim \mathcal{X}$.
    \item Without loss of generality we select, in turn, each feature $l_i$ in $\mathcal{E}'$ and perform a fixed intervention of $\pm 1$ 
    to obtain $\hat{l} = \Phi_a(\Phi_d (\Phi_e(z))), z \sim \mathcal{N}(\Phi_f (\Phi_g (l, do(l_i = \pm1))), 0.1)$; we then find, in $\hat{l}$, all other features affected by this intervention and note the change in their value with respect to their original value in $l$.
    \item Once we establish the change in value for feature $\hat{l_j}$ with respect to $l_j$ as a result of intervention on $l_i$, we perform a controlled intervention on feature $l_j$ with the observed change $\hat{l_j}$ resulting in $\hat{l'} = \Phi_a(\Phi_d (\Phi_e(z))), z \sim \mathcal{N}(\Phi_f (\Phi_g (l, do(l_j = \hat{l_j}))), 0.1)$, and note changes in its descendent feature values with respect to $\hat{l}$.
    \item We repeat the above two steps until all the features are  covered; if the relative change before and after an intervention is greater than a given threshold in an expectational sense (see  Equation ~\ref{eqn:edgeweightage}), we establish an edge between (nodes corresponding to) those two features ($l_i \rightarrow l_j$). Equation \ref{eqn:explicitedge} describes this process mathematically.
\end{enumerate}


An edge exists between nodes $l_i$ and $l_j$ only if there is a difference between $l_j$ and $\hat{l}_j$ upon intervention on $l_i$, conditioned on $\textbf{pa}_{l_i}$ (parent features of $l_i$).
Let us consider an ICP (see Definition~\ref{prp:causalstructure}) example, where $\node{1} \rightarrow \node{2} \rightarrow \node{3}$ and the second step establishes  causal relations $\node{1} \rightarrow \node{2}$ and $\node{1} \rightarrow \node{3}$, with respect to some threshold $T$ and ${v}^1_2, {v}^1_3$ corresponding to new values of $\node{2},\node{3}$, respectively, due to an intervention on $\node{1}$. 
In the third step, when we perform an intervention on $\node{2}$ by setting its value to ${v}^1_2$, let the observed value of $\node{3}$ be ${v}^2_3$; then, if $|{v}^2_3 - {v}^1_3| < \epsilon$ we establish the correct edge $\node{2} \rightarrow \node{3}$ by removing the spurious edge $\node{1} \rightarrow \node{3}$. 
In the case of loops, we use the edge-weight described in Equation \ref{eqn:edgeweightage} to determine the prominent causal direction.
Mathematically the formulation is described in Equation \ref{eqn:explicitedge}, where $\mathbb{I}$ is an indicator function determining the existence of an edge between $l_i$ and $l_j$.


\begin{equation}
    l_i \rightarrow l_j \triangleq \mathbb{I} [ \mathbb{E}_{l \sim \mathcal{E}'}(EW (\Phi_a(\Phi_d(\Phi_e(z)))_i, l_j ) | \textbf{pa}_{\vec{z}_i}) \!\!> \!\!T], 
    \; \; \; z \!\sim\! \mathcal{N}(\Phi_f (\Phi_g (l, do(l_i \!=\! l_i))), 0.1)
    \label{eqn:explicitedge}
\end{equation}

If the features are disentangled, intervening on $l_i$ should not affect $l_j$: we observe a similar effect in our experiments, which we discuss in the later section of the paper \ref{sec:results}.

\textbf{Graph correctness: }
The generated graph should be consistent and stable; we define correctness measures depending on these two factors: (i) stability captures the variation of a generated graph when the method is applied to different subsets of a dataset, while (ii) consistency captures the variation in the generated graph when the method is applied to the same data multiple times. 
We use both stability and consistency properties on the known subgraph for a given set of context features to define the graph correctness metrics.

\begin{asu}
\label{assumption:graph_correctness}
The correctness of the entire generated graph is proportional to the correctness of the subgraph with observed context features. We quantify the correctness of the subgraph by comparing it against the known ground-truth subgraph, along with stability and consistency properties.
\end{asu}

\textit{Remark:} As $|\mathcal{E}'_o| < |\mathcal{E}'_u|$ generally, a direct way to validate a generated graph would be via visual inspection, which may prove to be challenging in case of large graphs; in that case, we can use the subgraph of observed features and compare against ground-truth.

We sample $P$ random subsets with repetition from the test dataset and run $Q$ iterations of graph generation on each set.
The consistent graph generation behaviour in all $Q$ iterations measures the consistency of our method, while similar graph generation behaviour in all $P$ subsets of dataset measures the stability.
To quantify the correctness, we simply compare the edges in the generated subgraph with the known ground-truth graph and consider the average over all $PQ$ graphs, which can formally be described as: \textit{correctnessIndex} $ \triangleq \frac{1}{PQ} \sum_
P\sum_
Q\frac{\#CorrectEdgesPredicted - \#AdditionalEdges}{\#TotalEdges}$, where $\#CorrectEdgesPredicted, \#AdditionalEdges, \#TotalEdges$ correspond to the total number of correct edges predicted in a subgraph, wrong edges predicted in a subgraph and total number of ground-truth edges, respectively.
An edge with a wrong direction is considered an additional edge, so the defined metric accounts for both wrong directions and additional edges. 

\subsection{Explanations}
\label{subsec:exp}
The latent space feature vocabulary is much richer for extracting explanations beyond feature attributions and saliency maps. 
In contrast to importance scores and attention maps in input space, explanations based on latent features may help us analyse the model perception on input features.
Based on this, we generate globally-inspired local explanations, using the feature interaction graph (extracted as described in the previous section) as a global form of explanation. 
This feature interaction graph explains how the classifier perceives the relationships between various semantically meaningful concepts, which can reveal biases and be used to debug the classifier.
For obtaining local explanations, we follow the LIME \cite{lime} feature attribution method on the aligned latent features while preserving the feature interactions, indicating the significance of all the latent features in classifying an image into a specific class.
The generator model helps us visualize the effect of significant features and their interactions on a given image by constructing counterfactual samples.

We evaluate the local explanations with two measures: stability and faithfulness, as defined next.

\begin{wrapfigure}{r}{0.59\textwidth}
  \begin{center}
    \includegraphics[width=0.58\textwidth]{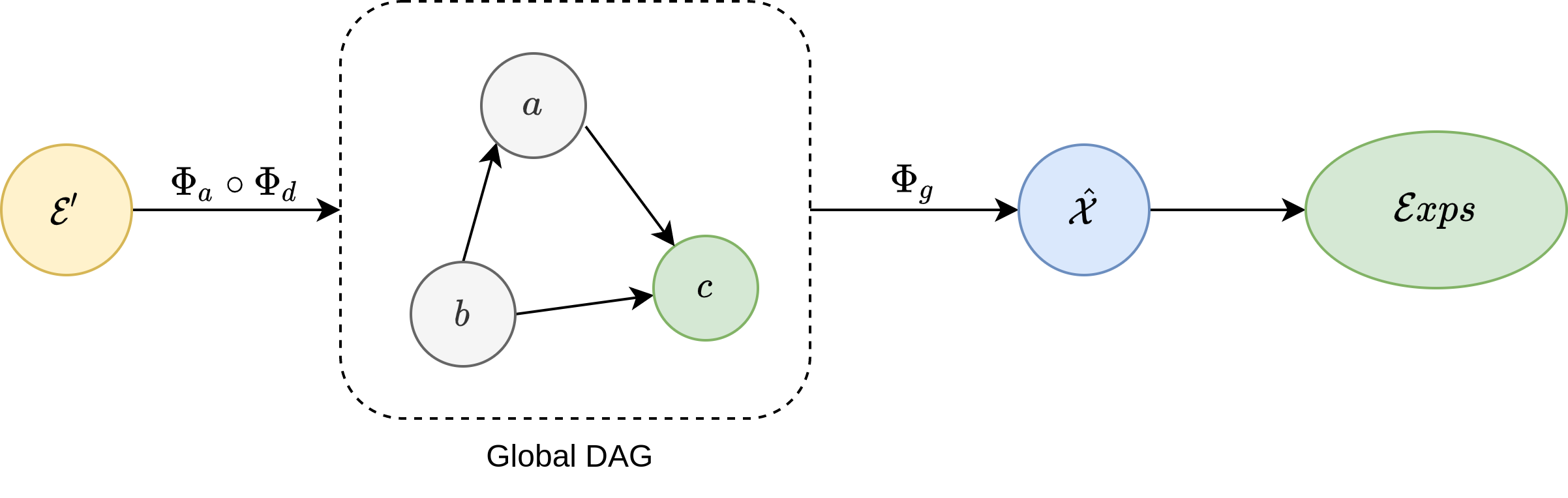}
    \caption{The flow of information in various blocks of our proposed framework from classifier features to explanations}
    \label{fig:framworkdag}
  \end{center}
  \vspace{-10pt}
\end{wrapfigure}

\textbf{Stability: }
We consider explanations to be stable if 
they are consistent across multiple iterations for the same image.
To quantify stability, we perturb an image sample with Gaussian noise generating $P$ samples to obtain $Q$ local explanations, one for each sample. 
We then consider a negative average of the variance in all the local explanations as stability. 
Formally, $stability(\mathcal{E}xps) \triangleq -\frac{1}{P} \Sigma_P \mathbb{E}_{x \sim \mathcal{E}xps}((x - \mathbb{E}(x))^2)$, where $\mathcal{E}xps$ is set of $Q$ explanations for one of the $P$ samples. 
The negative sign makes the metric directly proportional to the stability of explanations.

\textbf{Faithfulness: }
As our method follows the twin-surrogate model, explanations are a function of both data and classifier. 
We characterize explanations to be faithful if the contribution of 
the classifier is higher than the contribution of data.
We follow an information-theoretical approach to measure the flow of information \cite{o2020generative} to quantify faithfulness. Proposition \ref{prp:faithfulness} provides a quantitative metric.

\begin{dfn}
\label{dfn:informationflow}
The \emph{information flow} between two independent sets of nodes $A$ and $B$ \cite{ay2008information} is: 
$$\mathcal{I}(A \rightarrow B) = \int_A \mathcal{P}(a) \int_B \mathcal{P}(b \mid do (a)) \log \frac{\mathcal{P}(b \mid do(a))}{\int_{a'} \mathcal{P}(a') \mathcal{P}(b \mid do(a'))da'}db da$$
where $do(v)$ represents an intervention that fixes the value of a variable to $v$ irrespectively of its parents and $\mathcal{P}$ is a probability distribution.
\end{dfn}


\begin{prp}
Based on the above Definition \ref{dfn:informationflow} and with the reference to framework DAG in Figure \ref{fig:framworkdag},
we show that the bounded mutual information between $\mathcal{E}'$ and $\mathcal{E}xps$ is the same as the information flow from the classifier to the generated explanations.
Due to this, we consider the normalised mutual information as the `faithfulness' metric, given by $\Big( faithfulness index = \frac{\mathcal{I}(\mathcal{E}'; \mathcal{E}xps)}{\sqrt{\mathcal{H}(\mathcal{E}xps)\mathcal{H}(\mathcal{E}')}}\Big)$, where $\mathcal{H}(.)$ corresponds to entropy.
(The proof and reasoning for this proposition are given in appendix \ref{section:appendix})
\label{prp:faithfulness}
\end{prp}

\section{Results}
\label{sec:results}
We evaluate the performance of our proposed framework for both causal discovery and explanations; we use classifiers trained on two different datasets with observed context features, namely Morpho-MNIST\cite{castro2019morpho} and FFHQ\cite{karras2019style}.
We compare our graph generation technique against two standard methods for causal discovery, Linear Non-Gaussian Acyclic Model (LiNGAM) with latent confounders \cite{maeda2020rcd} and Greedy Equivalence Search (GES) \cite{chickering2002optimal}.
We compare our explanations against saliency-based methods, LIME \cite{ribeiro2016should}, DeepSHAP \cite{SHAP}, deepLIFT \cite{deeplift}, and gradCAM \cite{selvaraju2017grad}.

\begin{figure}
  \centering
    \subfloat[]{\includegraphics[width=.40\textwidth]{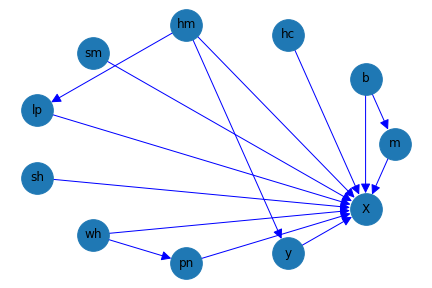}} \hfill
    \subfloat[]{\includegraphics[width=.28\textwidth]{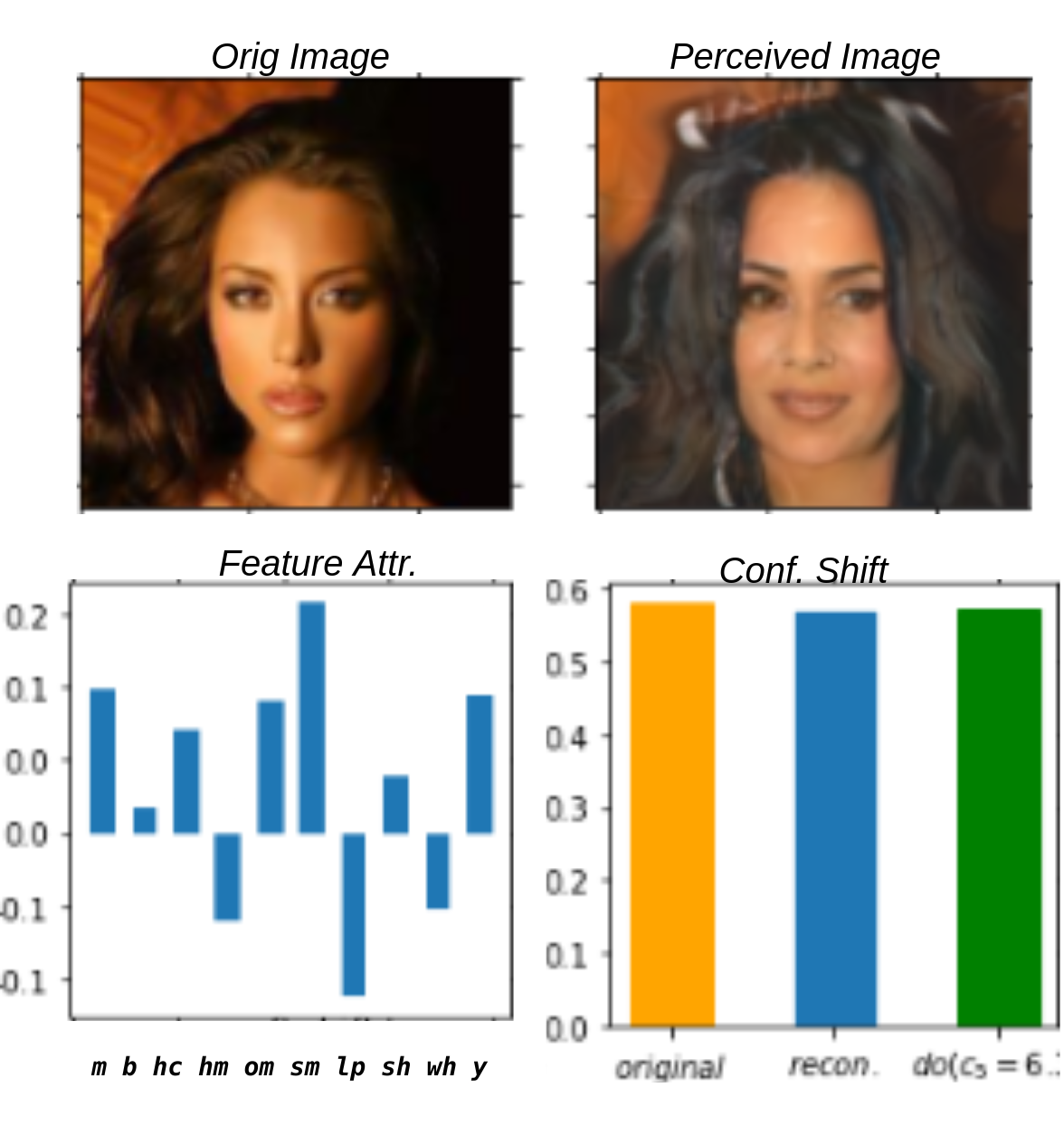}} \hfill
    \subfloat[]{\includegraphics[width=.31\textwidth]{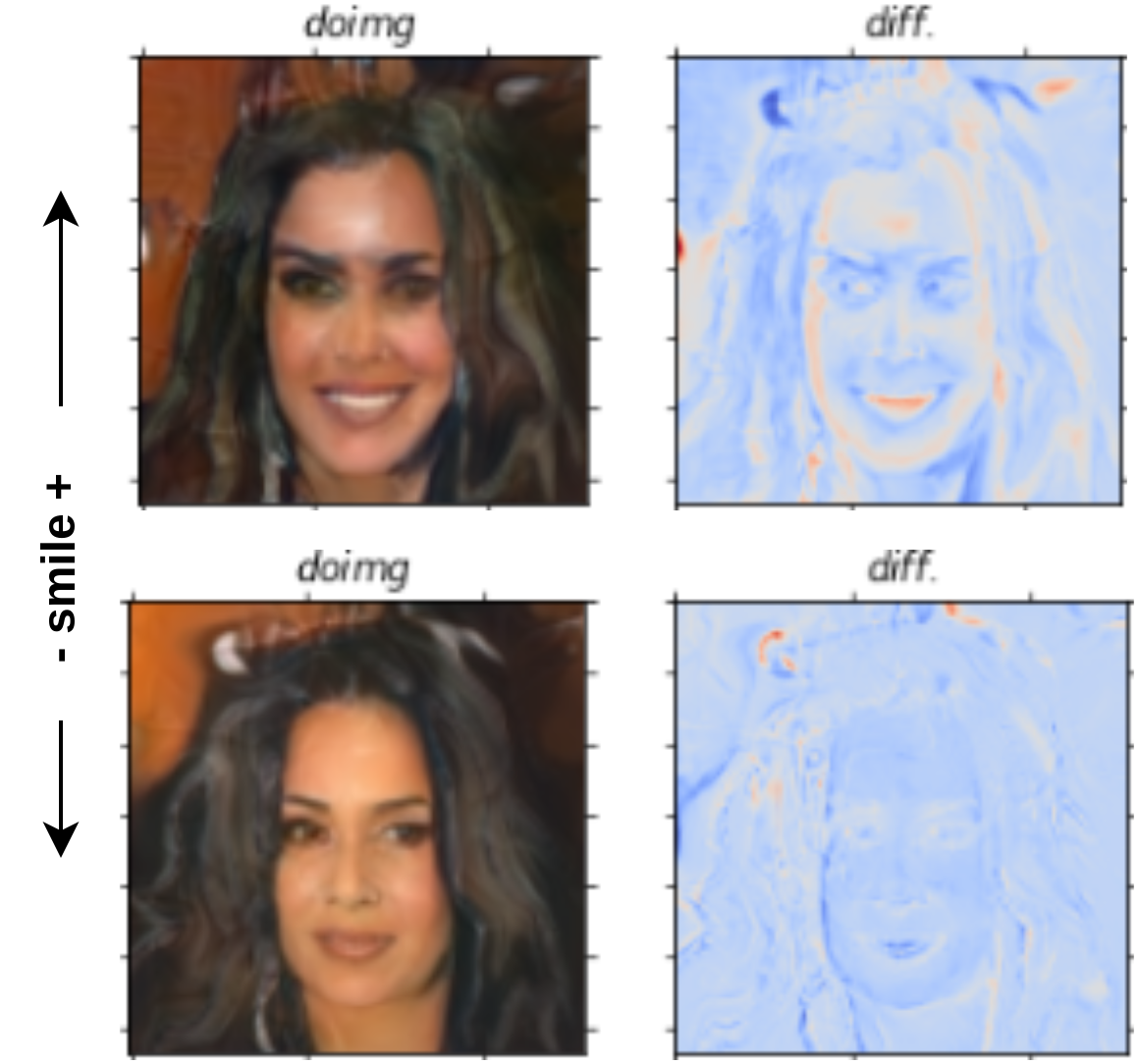}} \hfill
    \caption{(a) 
    DAG generated by our proposed method. (b) GLANCE explanations: the first row indicates the original image and the image perceived by the classifier
    ; the second row indicates feature importance scores and 
    effect on confidence scores with original, perceived, and intervened ( w.r.t the smile attribute) image
    . (c) The first column demonstrates an intervention on the perceived image and the second column corresponds to 
    the effect of an intervention.}
    \label{fig:ffhqresults}
    \vspace{-5pt}
\end{figure}

In the case of Morpho-MNIST, we tested our framework on four different synthetically generated datasets by varying causal relationships among features; all four data-generating processes are described in appendix section \ref{sec:casestudy1}.
Figure \ref{fig:mmnistresults} indicates a qualitative difference between our method and the other standard methods mentioned above. 
Existing explanations cannot understand the effect of intermediate features or cannot differentiate the effect of multiple features involved in making specific predictions. For example, these methods cannot differentiate between the effect of thickness and intensity or geometric features like ``loops'' in digits 8 and 6.
The global feature interaction graph generated 
by our method addresses this issue to an extent, as it captures complex feature interactions among aligned, semantically meaningful features.  
The feature interaction graph can further be used to investigate the locality of any given image or even used to generate 
counterfactual examples. Qualitative results are shown in Figures \ref{fig:mmnistresults} and \ref{fig:ffhqresults}.
To quantitatively compare explanations, we make use of the \textit{faithfulness index} 
and of the \textit{stability index} described in Section ~\ref{subsec:exp}. 
We report in Table \ref{table:explanationresults} the average faithfulness and 
stability indeces obtained for 1000 generated explanations
. 

In the case of the high resolution human faces dataset (FFHQ) \cite{karras2019style}, with an image size of 128x128, we explain the classifier trained to classify gender
.
For causal discovery, we only consider ten observed features out of forty given attributes in the dataset
,
selected based on the frequency of values of these features and their extent of being independent of each other (subjectively selected). 
A detailed list of selected features along with additional examples are described in appendix \ref{sec:casestudy2}.
Figure \ref{fig:ffhqresults}(a) describes the generated causal structure on ten observed context features, Figure \ref{fig:ffhqresults}(b) demonstrates a given image as perceived by the classifier along with importance scores for observed context features and 
the effect on confidence scores due to an intervention 
on the smile attribute, and Figure \ref{fig:ffhqresults}(c) describes the effect of an intervention on the smile attribute.

\begin{figure}
  \centering
    \subfloat[Ground-truth \\ sub graph]{\includegraphics[width=.25\textwidth]{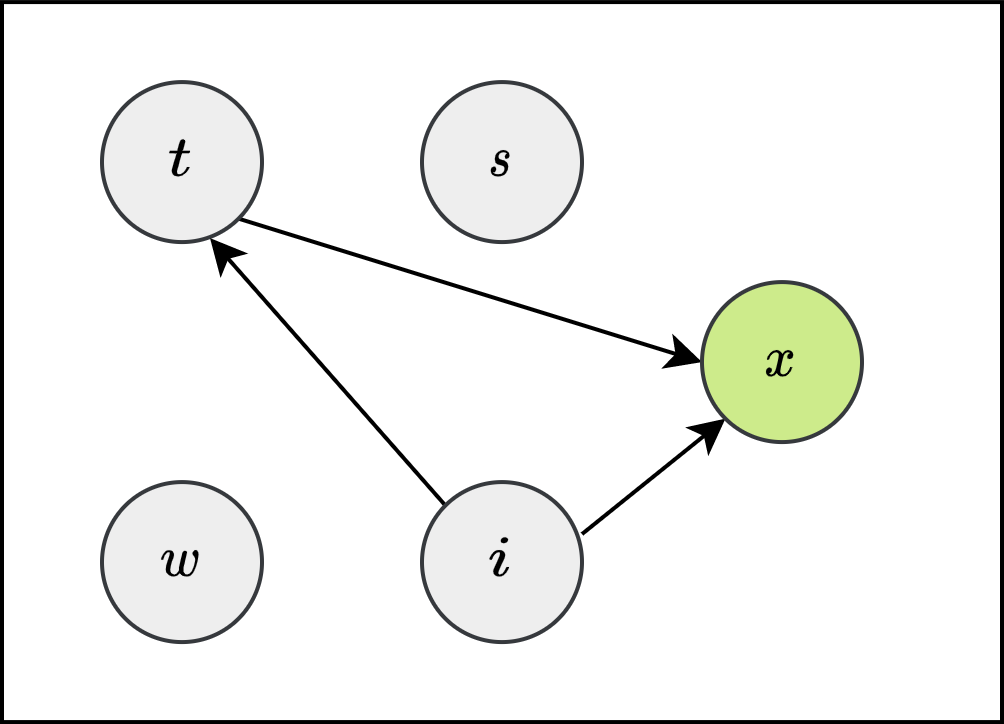}} \hfill
    \subfloat[GES discoveries \\ \textit{correctnessIndex=0.66}]{\includegraphics[width=.25\textwidth]{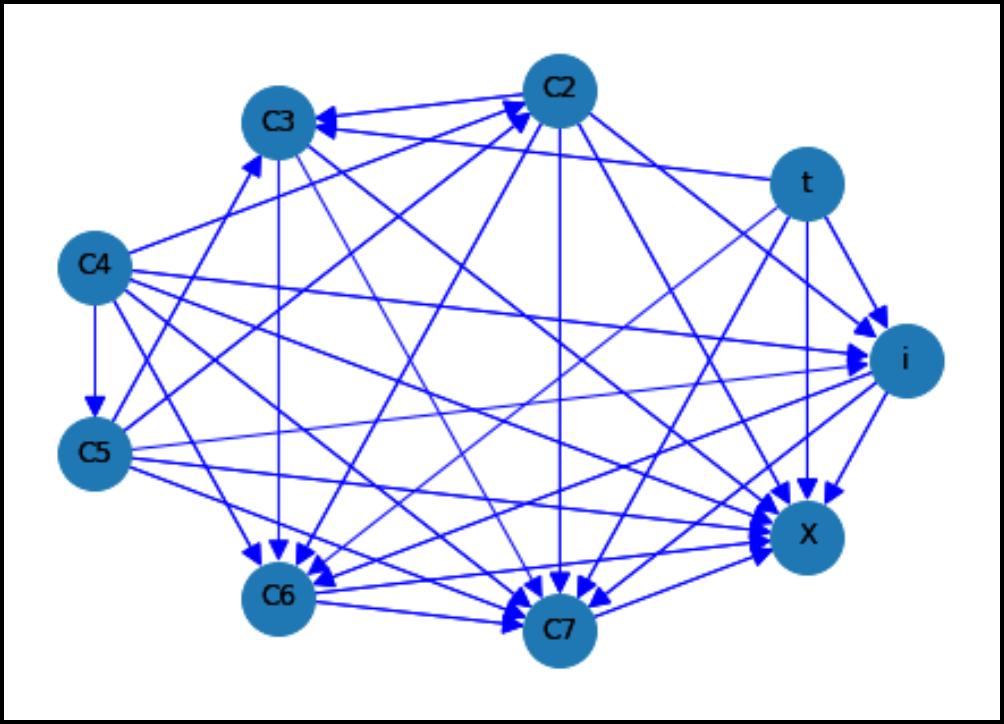}} \hfill
    \subfloat[LiNGAM discoveries \\ \textit{correctnessIndex=0.66}]{\includegraphics[width=.25\textwidth]{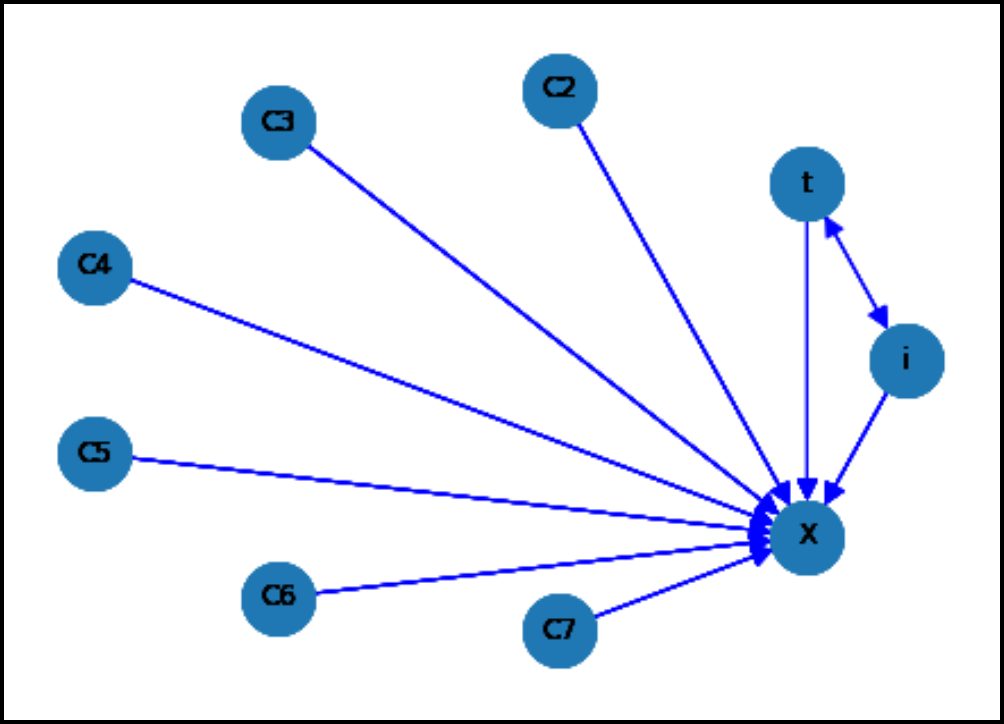}} \hfill
    \subfloat[Proposed \\ \textit{correctnessIndex=1.0}]{\includegraphics[width=.25\textwidth]{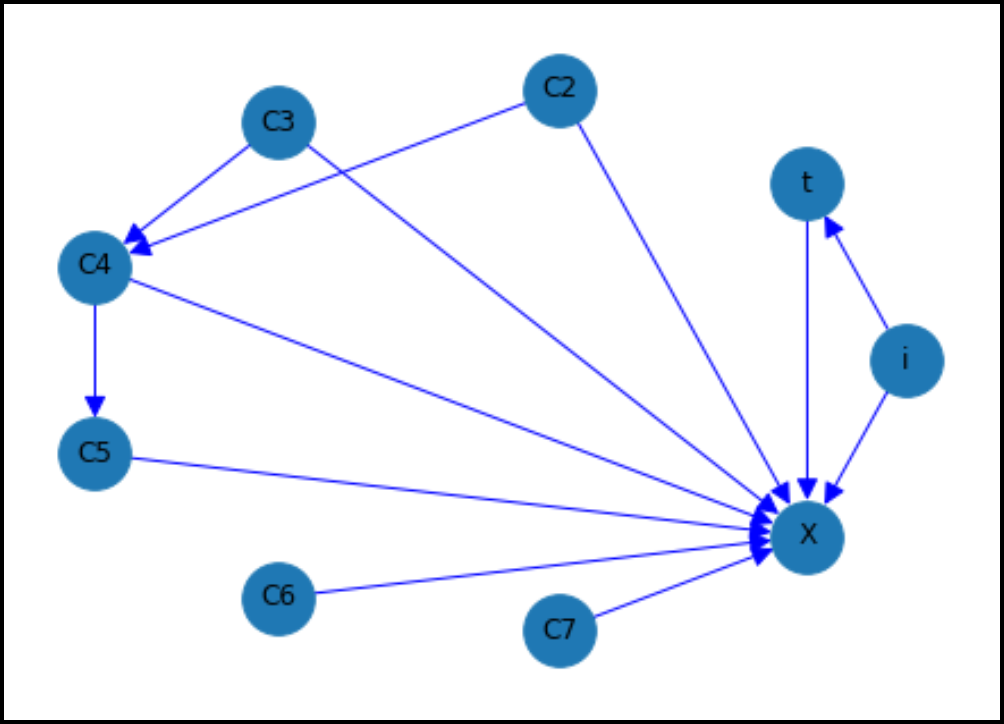}} \hfill
    \caption{(Quantitative and qualitative) Comparison of causal discovery methods
    .}
    \label{fig:causaldiscoveries}
\end{figure}

\begin{figure}
  \centering
    \subfloat[]{\includegraphics[width=.32\textwidth]{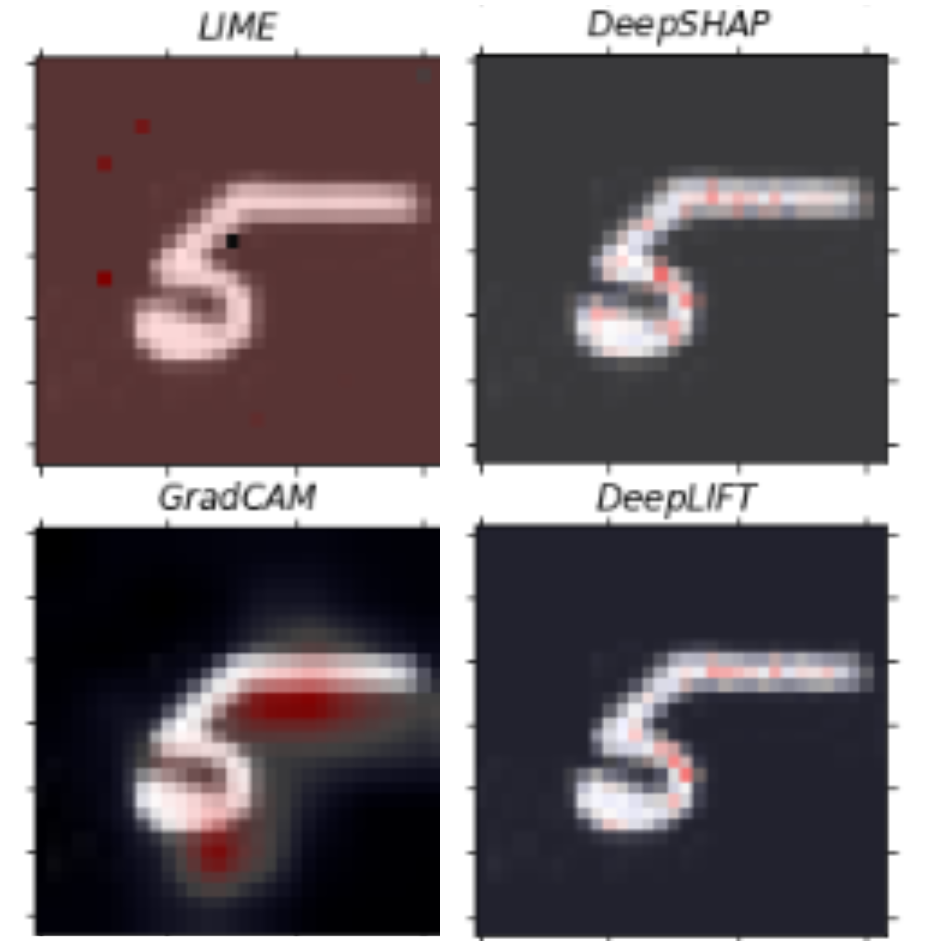}} \hfill
    \subfloat[]{\includegraphics[width=.30\textwidth]{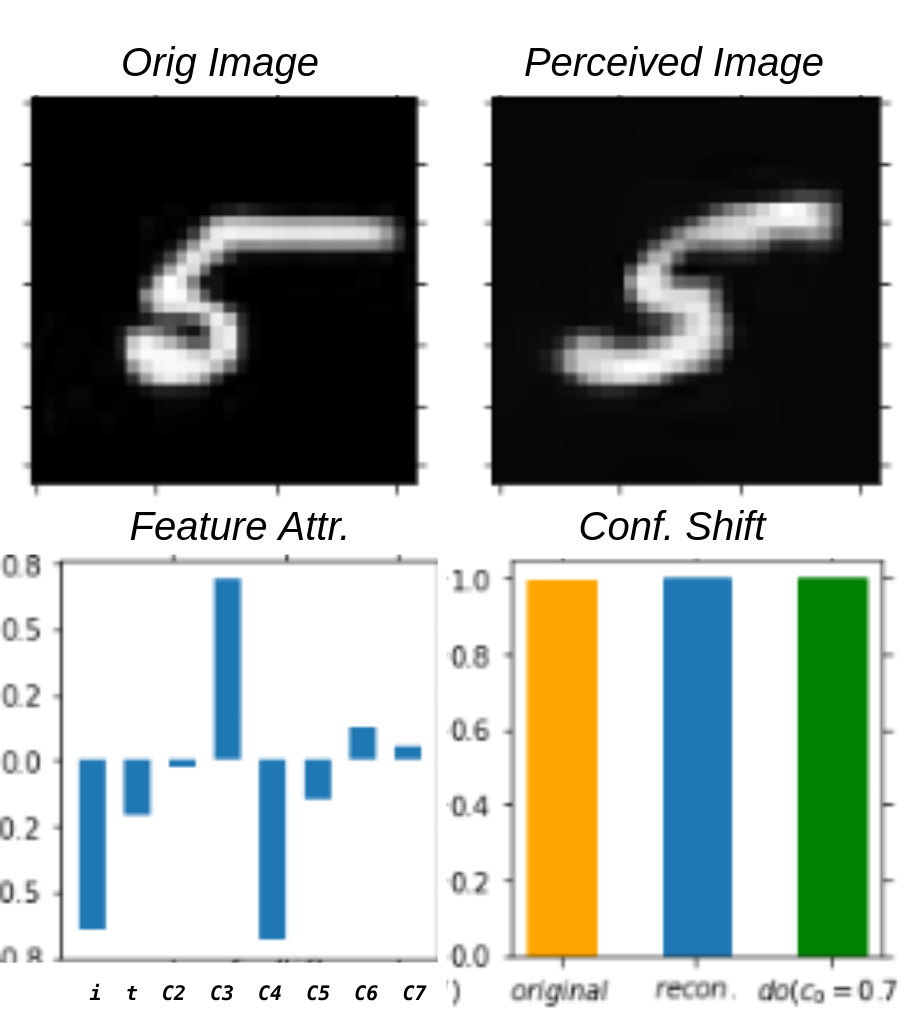}} \hfill
    \subfloat[]{\includegraphics[width=.35\textwidth]{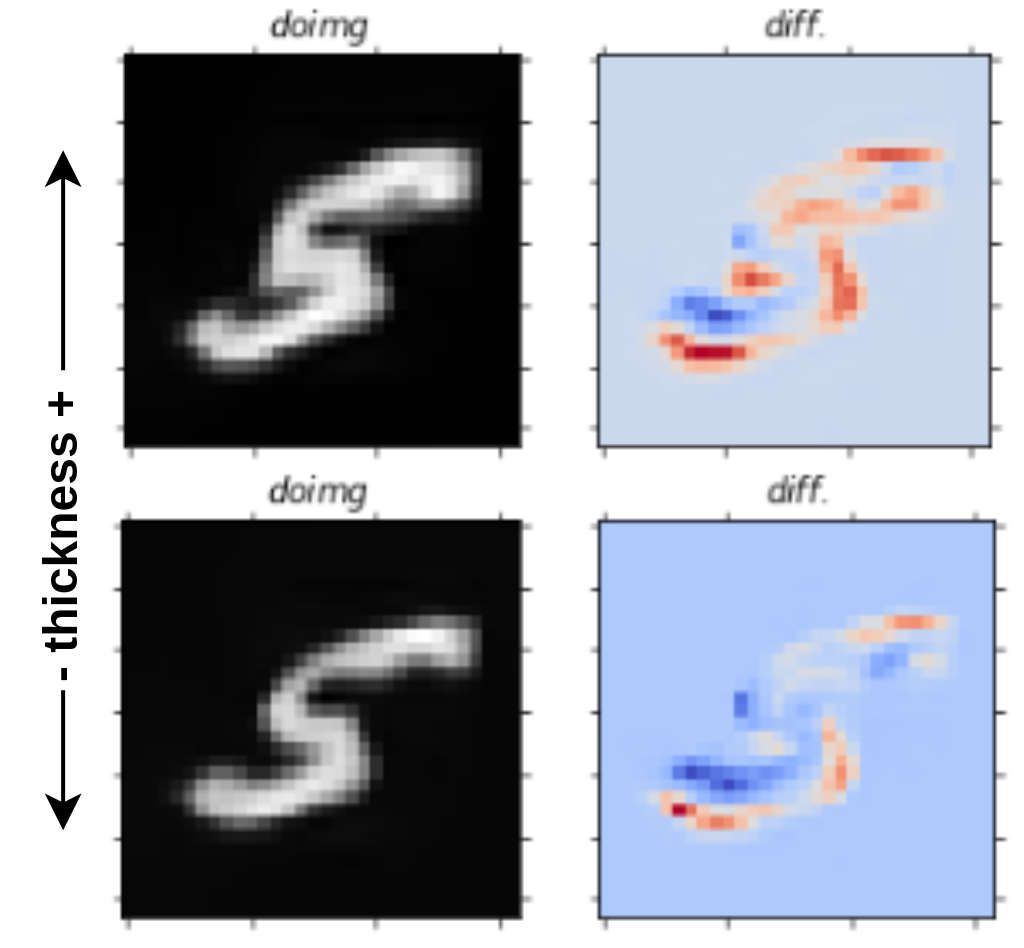}} \hfill
    \caption{
    Demonstration of the effectiveness of GLANCE explanations on the Morpho-MNIST-IT dataset (the global explanation graph is 
    in Figure ~\ref{fig:causaldiscoveries}(d)). (a) Demonstrates explanations obtained from four standard methods. (b) GLANCE explanations: the first row indicates the original image and the image perceived by the classifier
    ; the second row indicates aligned features' importance scores and 
    effect on confidence scores with original, perceived, and intervened (w.r.t the thickness attribute) images
    . (c) The first column 
    shows an intervention on the perceived image and the second column corresponds to 
    the effect of an intervention (difference between perceived and intervened images).}
    \label{fig:mmnistresults}
\end{figure}

\begin{table}[]
\caption{Quantitative comparison between multiple explanation methods with respect to faithfulness and stability properties (higher values indicate better performance).}
\vspace{10pt}
\centering
\begin{tabular}{@{}lccccc@{}}
\toprule
\begin{tabular}[c]{@{}l@{}}Metrics $\downarrow$ | Methods $\rightarrow$\end{tabular} & LIME & DeepSHAP & DeepLIFT & GradCAM & Ours \\ \midrule
\begin{tabular}[c]{@{}l@{}} \textit{Faithfullness Index}\end{tabular} &   0.22  &  0.67    &  0.92   &  0.22  &  \textbf{0.97}  \\
\begin{tabular}[c]{@{}l@{}} \textit{Stability Index}\end{tabular}  &      -1.40 &  -0.07  &  -0.04  & -2.53  &  \textbf{-0.02} \\ \bottomrule
\end{tabular}
\label{table:explanationresults}
\vspace{-10pt}
\end{table}

\section{Conclusion}
We present GLANCE, a novel explanation framework that uses latent space vocabulary to generate 
global explanations in terms of graphs and 
local explanations in terms of feature importance scores; 
then, a
generator can be used to visualize the effect of feature importance and interactions.
We validate both causal discovery and GLANCE explanations both qualitatively and quantitatively against existing standard explanations methods.
The proposed method for extraction of global 
explanations (in the form of DAGs) follows carefully constructed steps using the ideas of intervention, indicating the causal interaction and influence among features in latent space.
The quantification of faithfulness helps us consider explanations 
more carefully, and this helps us differentiate between explanations obtained from the underlying classifier model and explanations generated from data alone.
We do not consider assigning semantic meaning to aligned features in this work, which could be a possible direction for future work. 
 GLANCE, along with assigned semantic meaning to latent features similar to \cite{ghorbani2019towards}, would effectively communicate the decision making parameters in classifiers to humans, thereby increasing the quality of explanations.
Furthermore, extending the framework to explain other  models beyond classifiers for visual data, such as time series, text, or tabular data, can broaden the impact of our framework.

\section*{Acknowledgements}
This work was supported by UKRI [grant number EP/S023356/1], in the UKRI Centre for Doctoral Training in Safe and Trusted AI.
\bibliographystyle{unsrt}
\bibliography{main.bib}

\section{Appendix}
\label{section:appendix}

\begin{prp}
Based on the definition \ref{dfn:informationflow}, we show the bounded mutual information (normalized mutual information) can be considered as a 'faithfulness' metric to quantify the classifier's contribution to generating explanations.
\end{prp}

Let us consider the feature attribution based method, probability of generating explanation $\mathcal{E}xp \in \mathcal{E}xps$ can be formally described by a conditional $\mathcal{P}(\mathcal{E}xp \mid l), l \in \mathcal{E}'$.

$$\mathcal{I}(\mathcal{E}' \rightarrow \mathcal{E}xps) = \int_l \mathcal{P}(l) \int_{\mathcal{E}xp} \mathcal{P}(\mathcal{E}xp \mid do (l)) \log \frac{\mathcal{P}(\mathcal{E}xp \mid do(l))}{\int_{l'} \mathcal{P}(l') \mathcal{P}(\mathcal{E}xp \mid do(l'))dl'}d\mathcal{E}xp dl$$ 

As integrals are applied over entire space, intervention can be replaced by conditionals which simplifies the above equation as:
$$\mathcal{I}(\mathcal{E}' \rightarrow \mathcal{E}xps) = \int_l \int_{\mathcal{E}xp} \mathcal{P}(l) \mathcal{P}(\mathcal{E}xp \mid l) \log \frac{\mathcal{P}(\mathcal{E}xp \mid l)}{\int_{l'} \mathcal{P}(l') \mathcal{P}(\mathcal{E}xp \mid l')dl'}d\mathcal{E}xp dl$$ 
$$\Rightarrow \int_l \int_{\mathcal{E}xp} \mathcal{P}(l, \mathcal{E}xp) \log \frac{\mathcal{P}(\mathcal{E}xp \mid l)}{\mathcal{P}(\mathcal{E}xp)}d\mathcal{E}xp dl$$ 
$$\Rightarrow \int_l \int_{\mathcal{E}xp} \mathcal{P}(l, \mathcal{E}xp) \log \frac{\mathcal{P}(\mathcal{E}xp, l)}{\mathcal{P}(\mathcal{E}xp)\mathcal{P}(l)}d\mathcal{E}xp dl$$ 

$$\therefore \mathbb{I}(\mathcal{E}' \rightarrow \mathcal{E}xps) = \mathcal{I}(\mathcal{E}'; \mathcal{E}xps) \propto \frac{\mathcal{I}(\mathcal{E}'; \mathcal{E}xps) \cite{strehl2002cluster}}{\sqrt{\mathcal{H}(\mathcal{E}xps)\mathcal{H}(\mathcal{E}')}}$$

In the case of a counterfactual method, explanations $\mathcal{E}xps$ are a part of input data ($\mathcal{E}xps \in \mathcal{X}$). 
Without loss of generality, we can apply the same metric to quantify the explanation's faithfulness to a classifier. 


\section{Case Study 1: Morpho-MNIST}
\label{sec:casestudy1}

\begin{figure}
    \centering
    \subfloat[]{\includegraphics[width=0.24\textwidth]{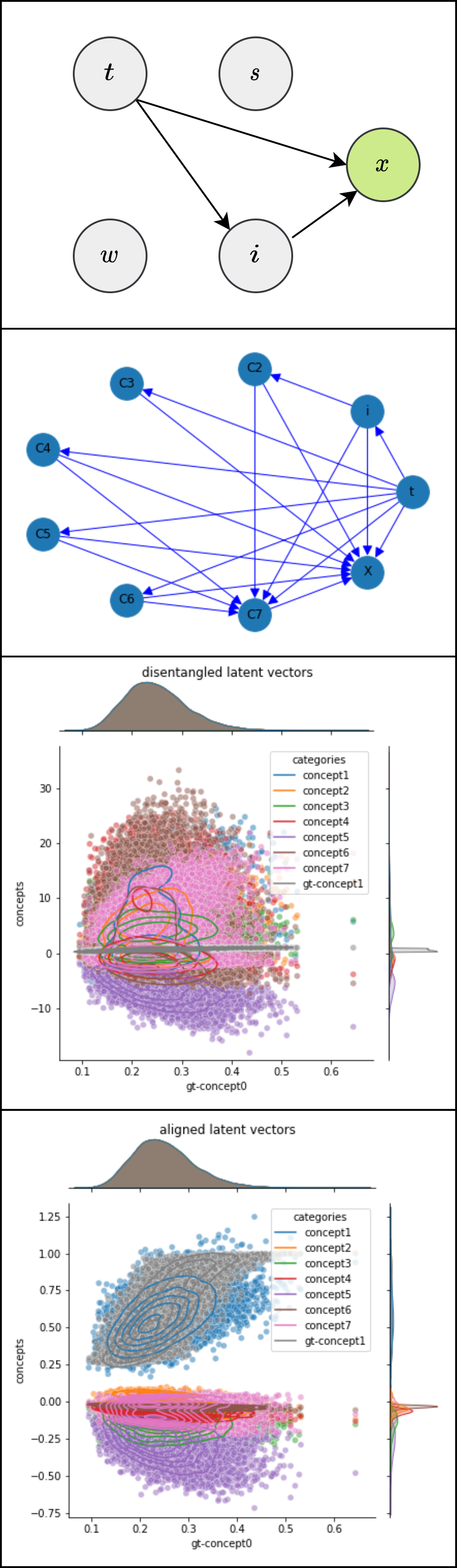}}\hfill
    \subfloat[]{\includegraphics[width=0.24\textwidth]{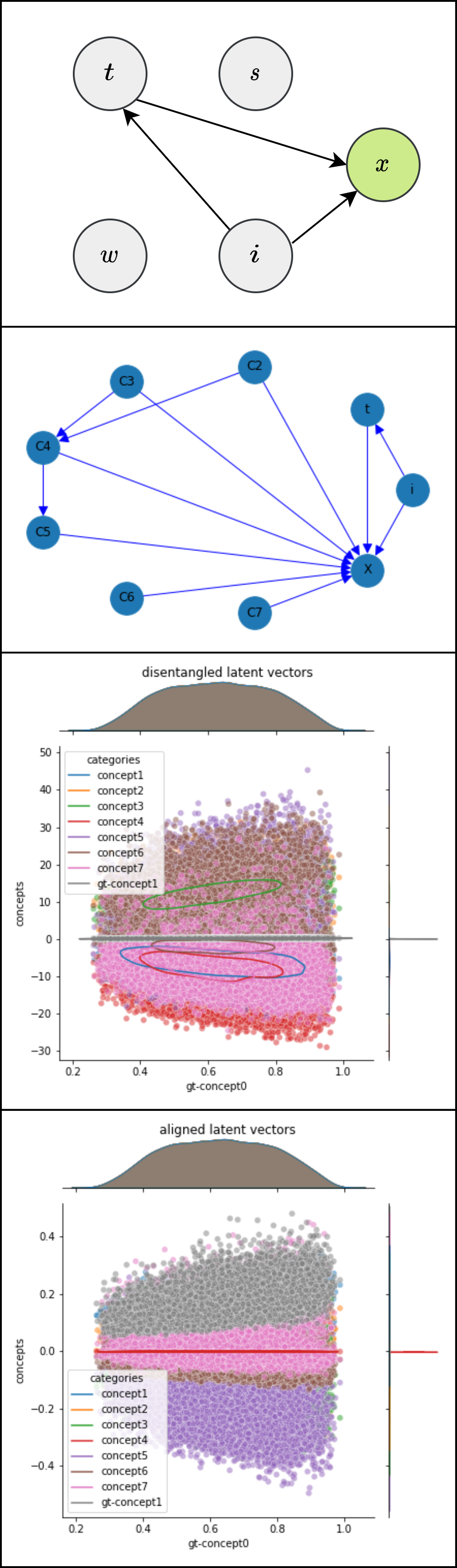}}\hfill
    \subfloat[]{\includegraphics[width=0.24\textwidth]{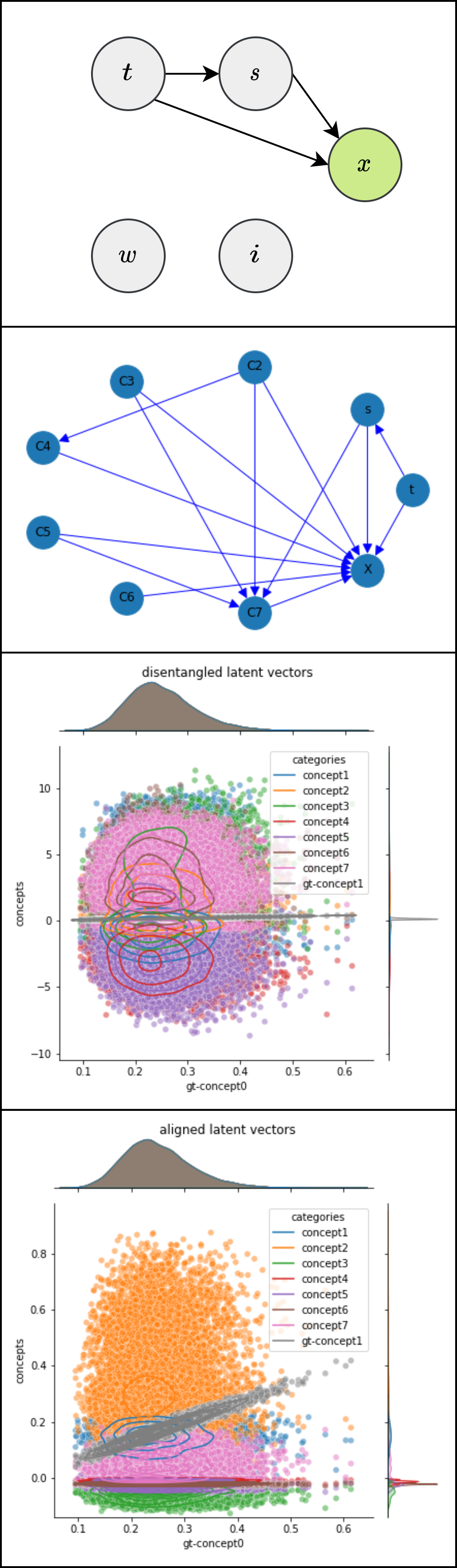}}\hfill
    \subfloat[]{\includegraphics[width=0.24\textwidth]{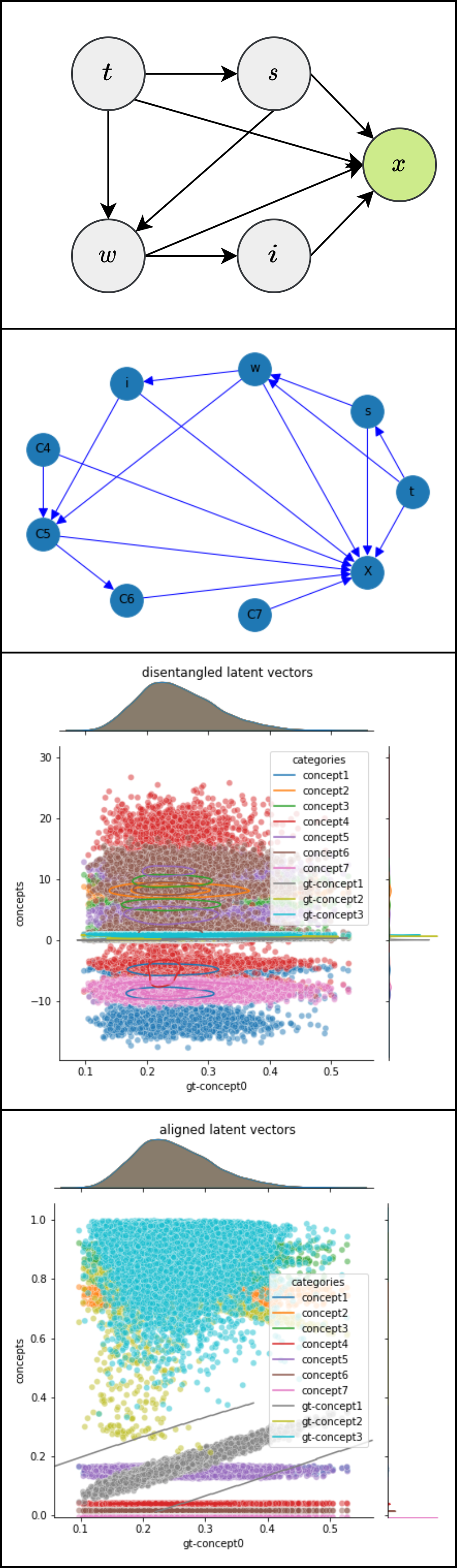}}\\
    \caption{Causal discoveries on various different data-generating processes. Top row describes the causal relationships followed in data-generating process, second row shows the discoveries made by our proposed method, third row shows the cluster formed by a feature disentanglement block to describe an alignment effect, fourth row describes the effectiveness of our alignment block in aligning model latent features to observed context features. 
    In (a) thickness causes intensity and both thickness and intensity causally affects image, the same behaviour can be observed in generated causal graph with graph \textit{correctnessIndex=1.0}.
    In (b) intensity causes thickness, we selected this example to examine algorithms behaviour in the case of a reversed causal link(w.r.t (a)), and the correct behaviour is observed in the generated graph with \textit{correctnessIndex=1.0}.
    In (c) thickness causes slant, again the generated graph shows similar behaviour with \textit{correctnessIndex=0.98}.
    In (d) we tried adding multiple causal variables with higher relationship depth, and our proposed method was able to reconstruct these relationships with \textit{correctnessIndex=0.94}.}
    \label{fig:synthDataExp}
\end{figure}

Here, we consider explaining a model trained on synthetic data based on MNIST digits \cite{castro2019morpho}. 
We define multiple data-generating process with four different variables thickness, width, slant, and intensity, and observe how our proposed method retrieves this causal structure using latent information via controlled interventions.
In this setup thickness corresponds to the stroke thickness of a digit, width corresponds to the total width of a written digit, slant corresponds to the shear factor along a horizontal direction, and intensity corresponds to the average intensity of pixels in a digit.
Functions $SetIntensity( · ; i)$, $SetSlant( · ; s)$, $SetWidth( · ; w)$, and $SetThickness( · ; t)$ refer to the operations applied to original MNIST digit to generate new image $x$ with desired properties by controlling image morphology.
Below we formally define 4 different data-generating senarios, Figure \ref{fig:synthDataExp} pictorially demonstrates causal structure used in data-generating performance and our model performance.

\textbf{Morpho-MNIST-TI}: In this setting we consider two causal variables thickness and intensity, where thickness causes intensity. 
Mathematically the functional relationship between variables are defined as described in equation \ref{eqn:morphoMNISTTIEqns}.

\begin{equation}
    \begin{aligned}
        &t := f_t \triangleq 0.5 + \epsilon_t \quad \epsilon_t \sim \Gamma(10, 5) \\
        &i := f_i \triangleq 64 + 191*\sigma(2*w + 5) + \epsilon_i \quad \epsilon_i \sim \mathbb{N}(0, 1)\\
        &x := f_x = SetIntensity(SetThickness(X; t); i)
    \end{aligned}
    \label{eqn:morphoMNISTTIEqns}
    \centering
\end{equation}

\textbf{Morpho-MNIST-IT}: In this experiment we inverted a directionality from previous setting resulting in intensity to cause thickness, which is mathematically described in equation \ref{eqn:morphoMNISTITEqns}

\begin{equation}
    \begin{aligned}
        &i := f_i \triangleq \epsilon_i \quad \epsilon_i \sim \mathbb{U}(60, 255)\\
        &t := f_t \triangleq 3 + \sigma(i/255) + \epsilon_s \quad \epsilon_s \sim \mathbb{N}(0, 0.5)\\
        &x := f_x = SetThickness(SetIntensity(X; i); t)
    \end{aligned}
    \label{eqn:morphoMNISTITEqns}
    \centering
\end{equation}

\textbf{Morpho-MNIST-TS}: In this setup we use thickness and slant as causal attributes, where thickness causes digit slantness, which is formally described in equation \ref{eqn:morphoMNISTTSEqns}

\begin{equation}
    \begin{aligned}
        &t := f_t \triangleq \epsilon_t \quad \epsilon_t \sim \Gamma(0, 5) \\
        &s := f_s \triangleq 10 + 5*\sigma(2*t - 5) + \epsilon_s \quad \epsilon_s \sim \mathbb{N}(0, 0.5)\\
        &x := f_x = SetSlant(SetThickness(X; t); s)
    \end{aligned}
    \label{eqn:morphoMNISTTSEqns}
    \centering
\end{equation}

\textbf{Morpho-MNIST-TSWI}: In this setup we increased a complexity by using intensity, thickness, slant, and digit width as a causal attributes, where thickness causes slant, thickness and slant causes width, and width causes intensity.
This data-generating process is formally described in equation \ref{eqn:morphoMNISTTSWIEqns}

\begin{equation}
    \begin{aligned}
        &t := f_t \triangleq \epsilon_t \quad \epsilon_t \sim \Gamma(0, 5) \\
        &s := f_s \triangleq 10 + 20*t + \epsilon_s \quad \epsilon_s \sim \mathbb{N}(0, 5)\\
        &w := f_w \triangleq  10 + 15*\sigma(0.5*t) - 0.25*s  + \epsilon_w \quad \epsilon_w \sim \mathcal{N}(0, 1) \\
        &i := f_i \triangleq 64 + 191*\sigma(w/25) + \epsilon_i \quad \epsilon_i \sim \mathbb{N}(0, 1)\\
        &x := f_x = SetIntensity(SetWidth(SetSlant(SetThickness(X; t); s); w); i)
    \end{aligned}
    \label{eqn:morphoMNISTTSWIEqns}
    \centering
\end{equation}



Figure \ref{fig:synthDataExp} row 2 describes the explicit graph generated as a result of our framework. 
In all the cases subgraph with nodes $\in \{t, i, w, s, x\}$ matches precisely with the causal structure followed in our data-generating process, with a graph \textit{correctnessIndex} close to 1.0.
This indicates the existence of implicit mechanisms and causal structures, providing global explanations for a given data-generating process.

\begin{figure}
  \centering
    \subfloat[First row shows an original image, second row describes reconstructed images, while third shows the effect of intervention, followed by difference between intervention and perceived images, and feature importance as perceived by classifier.
    Last row corresponds to feature attribution explanations on extracted features in latent space.]{\includegraphics[width=1.\textwidth]{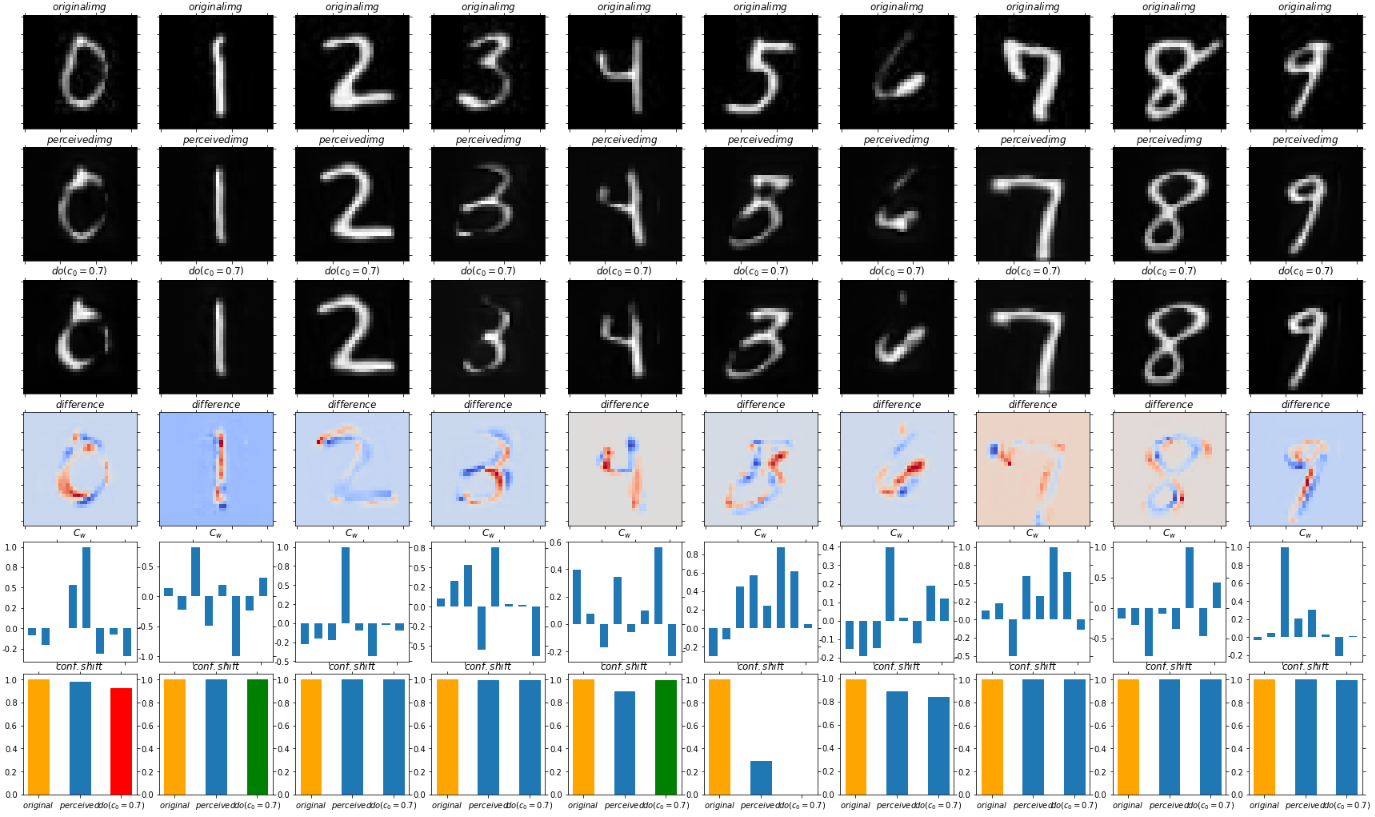}} \hfill
    \subfloat[Explanations generated using other standard methods: first, second, third, and forth row correspond to LIME, GradCAM, DeepSHAP, and DeepLIFT explanations respectively.] {\includegraphics[width=1.\textwidth]{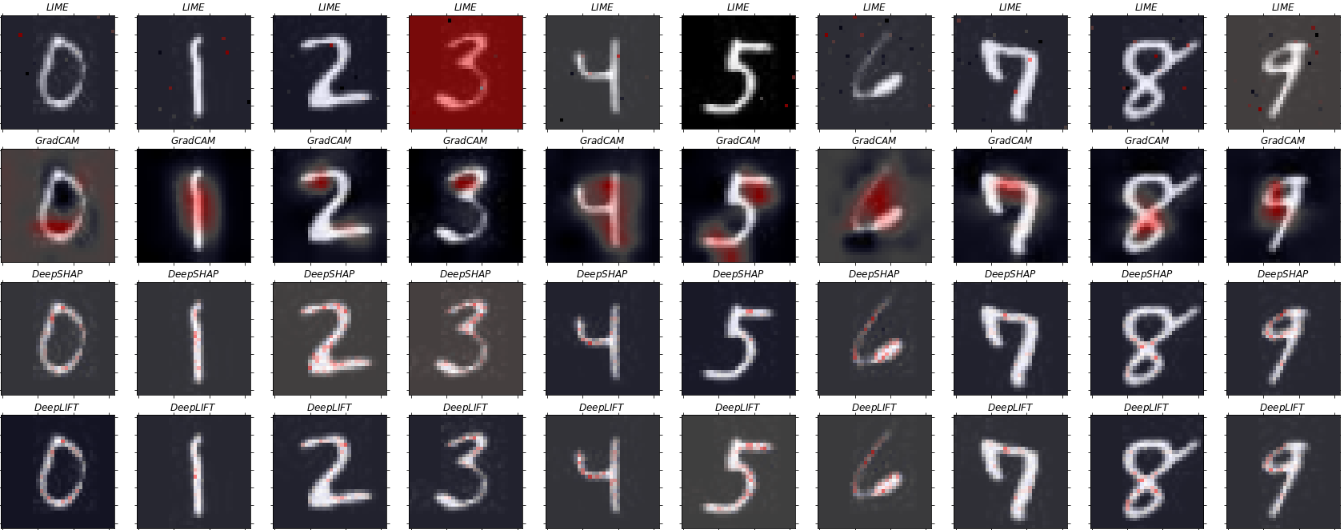}} \hfill
    
    \caption{Following figure demonstrates the effectiveness of our proposed explanations and other standard existing explanation frameworks}
    \label{fig:morhpoMNISTresults}
\end{figure}

To explain the importance of each feature for a classifier, we perform a LIME feature attribution while preserving the causal structure.
We perform a fixed interventional study on all the features by constructing multiple counterfactual images and observing the shift in confidence scores predicted by the classifier with respect to the original image, indicating the effect of features on the given classifier.
If confidence increases, we claim that a specific feature has a positive effect on a classifier; otherwise, it negatively affects a classifier.

As interventions may not have a monotonic effect, we conduct two specific queries, one positively increasing feature value while the other reducing the feature value.
Figure \ref{fig:morhpoMNISTresults} shows the generated counterfactual and the classifier probability describing the importance of a specific positively intervened feature.
Based on the extracted graph, an interventional behaviour of a feature on an image, and each feature's contribution to the final classifier's decision, we get a comprehensive idea of the classifier's reasoning.

\subsection{Comparative study}
\subsubsection{Graph Generation}
\begin{table}[]
\centering
\begin{tabular}{@{}llll@{}}

\toprule
\begin{tabular}[c]{@{}l@{}}Datasets $\downarrow$ \textbackslash\\ Methods $\rightarrow$\end{tabular} & \begin{tabular}[c]{@{}l@{}}LinGAM \\ Based \cite{maeda2020rcd}\end{tabular} & \begin{tabular}[c]{@{}l@{}}GES \\ Based \cite{chickering2002optimal}\end{tabular} & Ours \\ \midrule
Morpho-MNIST (TI)                                                                                &  0.84                                                                                &    0.66                                                                           &  \textbf{1.0}    \\
Morpho-MNIST (IT)                                                                                &  0.66                                                                              &  0.66                                                                              &    \textbf{1.0}  \\
Morpho-MNIST (TS)                                                                                & 0.82                                                                               &  0.66                                                                             &    \textbf{0.98}  \\
Morpho-MNIST (TSWI)                                                                              &  0.58                                                                             & 0.42                                                                              &  \textbf{0.94}    \\ \bottomrule
\end{tabular}
\vspace{5pt}
\caption{Table describes quantitative comparison between causal multiple causal discovery methods on four different versions of Morpho-MNIST dataset as described in \ref{fig:synthDataExp} using graph \textit{correctnessIndex}.}
\label{table:discoveryresults}
\end{table}


Most of the existing causal discoveries method try to extract relationships between nodes by assuming a particular structure of models. 
However, in our case, since we have access to an implicit causal model, we consider trained models as an oracle to perform specific interventions.
As we observe feature behavior against an actual cause rather than a model hypothesis, we have the flexibility to extract feature relations without any explicit assumptions.
We use the graph correctness index defined in \ref{subsec:graphs} to quantify the performance difference between all three methods; table \ref{table:discoveryresults} describes the results. 

\subsubsection{Explanation}



As previously mentioned, In this study we compare our method against standard saliency based explanation methods, we consider LIME \cite{ribeiro2016should}, DeepSHAP \cite{SHAP}, deepLIFT \cite{deeplift}, and gradCAM \cite{selvaraju2017grad} explanation and compare them against our explanations.
These methods generate an attention map for an input image given the model's prediction confidence on that image.
These explanations can provide a simple understanding of what the network is looking at in making certain decisions, but they fail to understand complex feature interactions or even fail to capture relations between pixels in input space.
These explanation methods do not yield a way to quantify the faithfulness of their generated explanations, which raises the question of trust in the explanations themselves.
Explanations generated using our method can overcome this kind of issue.
Figure \ref{fig:morhpoMNISTresults}, demonstrates our framework performance on multiple images, while Figure \ref{fig:synthDataExp} shows the behavior of intermediate latent features.

\begin{wrapfigure}{r}{0.4\textwidth}
  \begin{center}
    \includegraphics[width=0.39\textwidth]{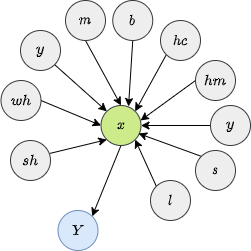}
    \caption{Assumed true causal structure between attributes and image.}
    \label{fig:ffhqgraph-gt}
  \end{center}
\end{wrapfigure}

\newpage
\section{Case Study 2: FFHQ}
\label{sec:casestudy2}

\begin{figure}
  \centering
  \includegraphics[width=1\textwidth]{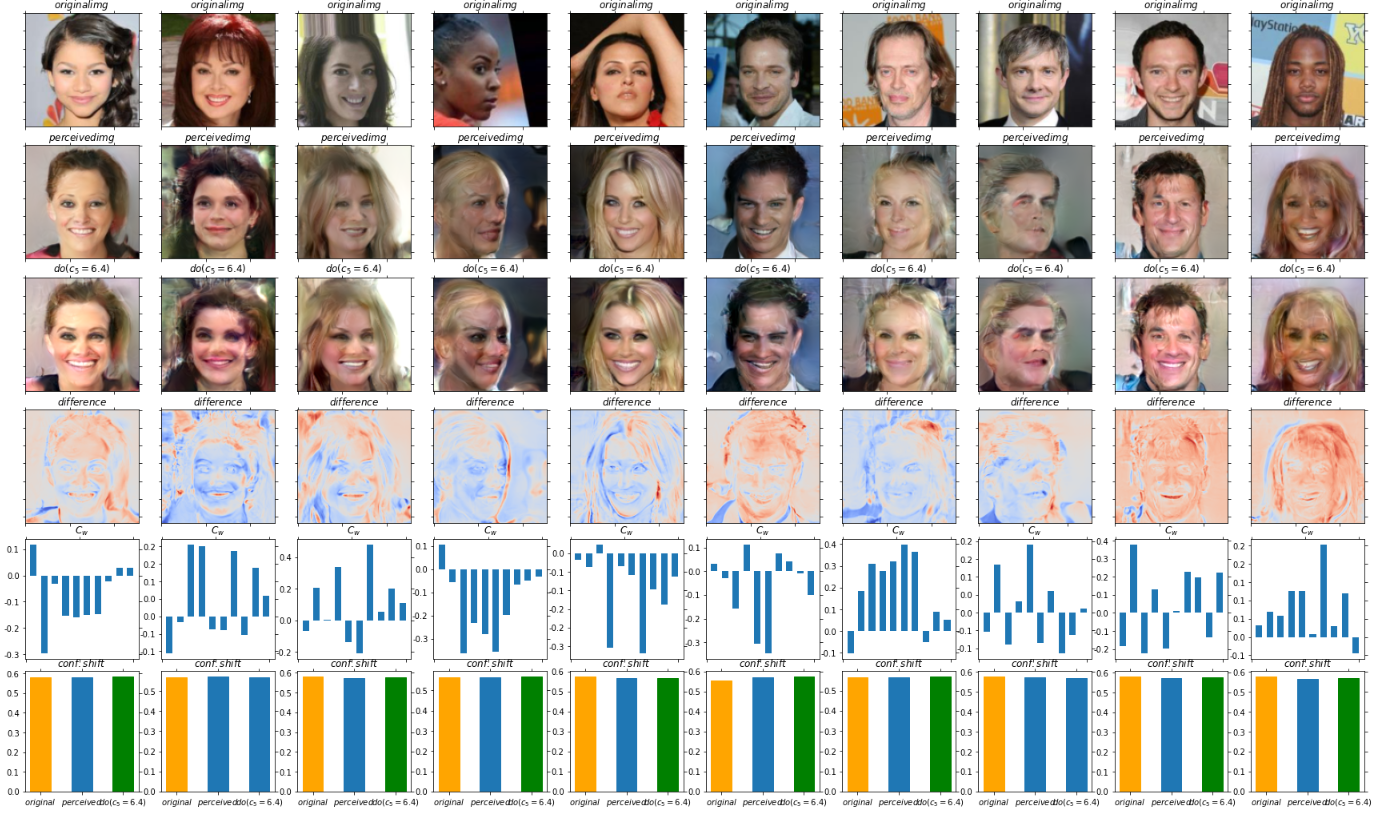}
    \caption{First row shows an original image, second row describes reconstructed images, while third shows the effect of intervention, followed by difference between interventional and perceived images, and feature importance as perceived by classifier.
    Last row corresponds to feature attribution explanations on extracted features in latent space}
    \label{fig:ffhqexplanations}
\end{figure}

For the second case study, we consider the high resolution human faces dataset (FFHQ) \cite{karras2019style}, this dataset consists of approximately 200k images of 128x128 resolution with 40 different binary attributes, and the task is to categorize images based on gender (0 = male; 1=female).
As the causal structure is unknown, we consider only ten significantly present attributes in the dataset.
In our experiment, these attributes are subjectively based on their interactions with respect to other attributes.
We pick features that are seemingly orthogonal to one another because that helps us to assume ground-truth causal structure to follow naive Bayes structure with all the selected features.
The ten attributes which we used in our experiment include (\emph{sh: straight-hair, wh: wavy-hair, y: young, m: mustache, b: beard, hc: high-cheekbones, hm: heavy-makeup, s: smiling, l: lipstick, o: open-mouth}), and the structure of assumed ground truth DAG is described in Figure \ref{fig:ffhqgraph-gt}.

For causal discovery and explanations, we consider 512 latent features to capture all the information in the data distribution. 
Obtained explanations from our method for the classifiers trained on this dataset are described in Figure \ref{fig:ffhqexplanations}. 
Here, we consider 'smile' as an interventional attribute; the higher attention around the mouth region can be easily seen in different images, indicating the effect of smile intervention. 
In this current work, we faced challenges with generating high quality counterfactuals. In the future, we are planning to extend this work with auxiliary modules to learn and associate causal attributes with generating high quality and meaningful counterfactuals.

\section{Training}
	
We trained all our models on a system with GPU: Nvidia Telsa T4 16GB, CPU: Intel(R) Xeon(R) Gold 6230, and RAM of 384GB.
In case of Morpho-MNIST, images were resized to $32\times32$ and models were trained with batchsize of 32 with learning rate = 1e-3, $\lambda_1 = 10., \lambda_2 = 30., \lambda_3 = 30.0, \& \lambda_4 = 1.0$.
In case of AFHQ, images were resized to $128\times128$ and models weere trained with batchsize of 8 with learning rate = 2e-4, $\lambda_1 = 10.0, \lambda_2 = 40.0, \lambda_3 = 80.0, \& \lambda_4 = 1.0$.

\end{document}